\documentclass[conference]{IEEEtran}
\IEEEoverridecommandlockouts
\usepackage{cite}
\usepackage{amsmath,amssymb,amsfonts}
\usepackage{algorithm,algorithmic}
\usepackage{graphicx}
\usepackage{textcomp}
\usepackage{xcolor}
\def\BibTeX{{\rm B\kern-.05em{\sc i\kern-.025em b}\kern-.08em
    T\kern-.1667em\lower.7ex\hbox{E}\kern-.125emX}}
\begin{document}

\title{Network Anomaly Detection Using Federated Learning}
\author{\IEEEauthorblockN{\textsuperscript{} William Marfo}
\IEEEauthorblockA{\textit{Department of Computer Science} \\
\textit{University of Texas at El Paso}\\
El Paso, USA \\
wmarfo@miners.utep.edu}
\and
\IEEEauthorblockN{\textsuperscript{} Deepak K. Tosh}
\IEEEauthorblockA{\textit{Department of Computer Science} \\
\textit{University of Texas at El Paso}\\
El Paso, USA \\
dktosh@utep.edu}

\and
\IEEEauthorblockN{\textsuperscript{} Shirley V. Moore}
\IEEEauthorblockA{\textit{Department of Computer Science} \\
\textit{University of Texas at El Paso}\\
El Paso, USA \\
svmoore@utep.edu}

}

\maketitle

\begin{abstract}
Due to the veracity and heterogeneity in network traffic, detecting anomalous events is challenging. The computational load on global servers is a significant challenge in terms of efficiency, accuracy, and scalability. Our primary motivation is to introduce a robust and scalable framework that enables efficient network anomaly detection. We address the issue of scalability and efficiency for network anomaly detection by leveraging federated learning, in which multiple participants train a global model jointly. Unlike centralized training architectures, federated learning does not require participants to upload their training data to the server, preventing attackers from exploiting the training data. Moreover, most prior works have focused on traditional centralized machine learning, making federated machine learning under-explored in network anomaly detection.
Therefore, we propose a deep neural network framework that could work on low to mid-end devices detecting network anomalies while checking if a request from a specific IP address is malicious or not.
Compared to multiple traditional centralized machine learning models, the deep neural federated model reduces training time overhead. The proposed method performs better than baseline machine learning techniques on the UNSW-NB15 data set as measured by experiments conducted with an accuracy of 97.21\% and a faster computation time.

\end{abstract}


\section{Introduction}
The internet is turning out to be an integral part of everyone's lives as more and more devices are being connected to serve societal needs. 
Our work is motivated by two major observations. Firstly, one drawback of connecting to the network is the threat of network attacks that can compromise users' private information, leading to data loss and adversely affecting productivity. There are several traditional security mechanisms to defend against these attacks, such as firewalls, virtual private networks (VPNs), demilitarized zones (DMZs), and vulnerability scanners. One way to prevent these attacks is early detection and prevention. However, these kinds of architecture do not scale very well because of their centralized nature. Secondly, we observe from heuristics and data set distributions that the majority of the requests made to a server are innocuous. Therefore, almost all server request data sets are highly imbalanced, weighted highly towards the harmless requests \cite{b1,b2}. 

To solve these issues, we propose a federated deep learning model that keeps track of session requests from different devices by using their IP addresses and port numbers to determine whether the request seems anomalous. Network anomaly detection using machine learning (ML) helps by analyzing different features of the request to verify if a request coming from a device is anomalous or not \cite{b3}. Further, as we use a federated approach, we have no single point of failure, making our approach scalable and more robust. 

With its scalability, federated learning (FL) helps tackle this problem. Since FL is not centralized, it removes dependence on a central server that plagues the scalability of large networks' cores. The workload is distributed by delegating the learning to edge servers from the singleton global server. Besides having the benefit of higher efficiency in the system, the ensemble of the independent edge servers also contributes to higher accuracy. We demonstrate both benefits in our results in Sec. IV.

Our FL architecture uses a hierarchical approach in which edge servers are added between clients and global servers to reduce the workload on the global server and the failure rates. In FL, only models with trainable weights parameters can be aggregated without a drop in their performance. For our network anomaly detection experiments, we propose using artificial neural networks (ANN) as it not only performs well in comparison to other models but is also an excellent candidate to be used in the federated learning environment due to their flexible nature \cite{survey}. 

The \textbf{contributions} of this paper can be summarized as follows.
\begin{itemize}

    \item We present a hierarchical federated learning architecture (HFL) to increase the scalability of FL for network anomaly detection. 
    
    \item We propose an HFL framework for network anomaly detection on the UNSW-NB15 dataset, which is the first on this dataset in literature to the best of our knowledge.\cite{b1}.
    
    \item We conduct and report comprehensive experiments to show the performance improvements of our federated learning model over the traditional centralized models that were trained individually on local data of clients. 
    
\end{itemize} 

The rest of this paper is organized as follows: Section 2 introduces background knowledge about federated learning and network anomaly detection. Section 3 details our proposed architecture for network anomaly detection using HFL. Experimental data and results are given in Section 4. In Section 5, we provide a summary and discuss future works.

\section{Background}
Network anomaly detection using FL is an application of ML in cybersecurity. This section discusses the concepts of federated learning, network anomaly detection, and some current work in these domains.

\subsection{Federated Learning}
Presently, the trend of using applications based on client-server architecture, such as cloud computing and web or mobile applications, is increasing \cite{survey}. Also, with technological advancement, the use and trust of artificial intelligence (AI)-based applications are increasing. FL is a valuable technique for performing AI-based tasks using ML or deep learning (DL) applications based on client-server architecture \cite{client}. In FL, ML/DL models are trained in a distributed and collaborative environment by ensuring the privacy of sensitive data of clients and the performance of the ML/DL models. In order to preserve the privacy of sensitive data and the model's performance, two types of models are used in federated learning, including local and global models. The client's data is not moved directly from one client's device to the server in federated learning. Instead, local models are trained on each client's local devices, and only the parameters of models trained on the client's devices are sent to a server, where the parameters of global models are updated using these parameters from the local models. In this way, not only is the user's privacy preserved, but it also causes less overhead on a server, reducing the computational resources and cost. Due to additional computational overhead on the global server, we can only add new devices to a limited scale. Also, vertical scaling is an expensive thing to do. Most of these issues are resolved using a hierarchical architecture in a federated learning environment. 

In \cite{comparison1}, the researchers used the same data set as ours and used ANN-based learning in a centralized fashion. The primary novelty of that work was feature selection. They achieved accuracy between 88.13 and 90.85\%. We show that our FL model outperforms this accuracy more efficiently and quickly. 

Moreover, FL methods have been proposed for intrusion detection in various applications, including large-scale cyber-physical systems, Wi-Fi networks, and others. Researchers in \cite{15} adopted a deep neural network to learn the hierarchical representations of private network data and achieved an accuracy of 99.27\% for intrusion detection with a focus on edge nodes and a cloud server.
With a suitable dataset, one can get the best precision for the ML techniques used in FL learning. The authors in \cite{16} introduced a study on several past methods applied for intrusion detection using FL learning. The datasets used had 5,874,010 trainable parameters where they utilized the framework to achieve better communication-computation trade-offs in a client-edge-cloud HFL system.

Authors in \cite{17,18} used an FL framework to train a deep learning model in IoT heterogeneous systems and reduced communication overhead by 75-85\%. In \cite{19}, researchers proposed an FL framework to optimize user assignment and allocate resources to provide a scalable production system based on Tensorflow.

In our HFL, edge servers are added as intermediate servers between clients and global servers. Edge servers are parameter servers that add more distribution to the FL architecture. Multiple clients are connected to a single-edge server based on availability. The edge server receives the parameters of local models from clients, performs aggregations, and passes these parameters to global servers. Due to this computational distribution, overhead on the global server is reduced. As edge servers are connected to a relatively smaller number of clients, edge servers are also more efficient. In case of failure of an edge server, a client can connect to other edge servers, which also increases the reliability of our FL architecture. 

As discussed in the introduction, this problem is biased in one class. Therefore, studies have been done in this context to tackle the class imbalance problem. In \cite{smote}, the authors used the same data set as ours and applied oversampling method (SMOTE) to handle the class imbalance problem, and used Random Forest (RF) classifier for the classification task and achieved 95.1\% accuracy. Our method not only aims to outperform this in accuracy but also makes a more robust model as FL is distributed while addressing the issue of class imbalance. 

\subsection{Network Anomaly Detection}
Devices connected to networks are vulnerable to attacks from other malicious devices. Network anomaly detection is used to identify requests from such malicious devices as anomalies considering that the requests from normal devices will have different characteristics than those from malicious devices. The idea is to have an early detection and prevention system that can identify if a request from a device deviates from the usual trend and deny the request if the algorithm identifies it as such. In ML, many techniques exist for anomaly detection based on available data, and classification algorithms can be used if data is labeled \cite{anomaly-survey1, anomaly-survey3, unsupervised}; however, unsupervised algorithms such as clustering or autoencoders are used in the case of unlabeled data \cite{unsupervised}. In our experiments, we use the labeled UNSW-NB15 dataset \cite{b1} to verify if a request is anomalous or not, and we tackle the problem as a classification task in an HFL environment.

\section{Proposed Approach}
Our DL-based ANN model is trained in an HFL environment to detect anomalous requests from different devices on the connected network. In this section, we discuss our approach in two parts. In the first part, we discuss the structure of our HFL environment as well as the architecture of the ANN model we used for anomaly detection in the second part.

\begin{figure}[h]
\centerline{\includegraphics[width=0.42\textwidth]{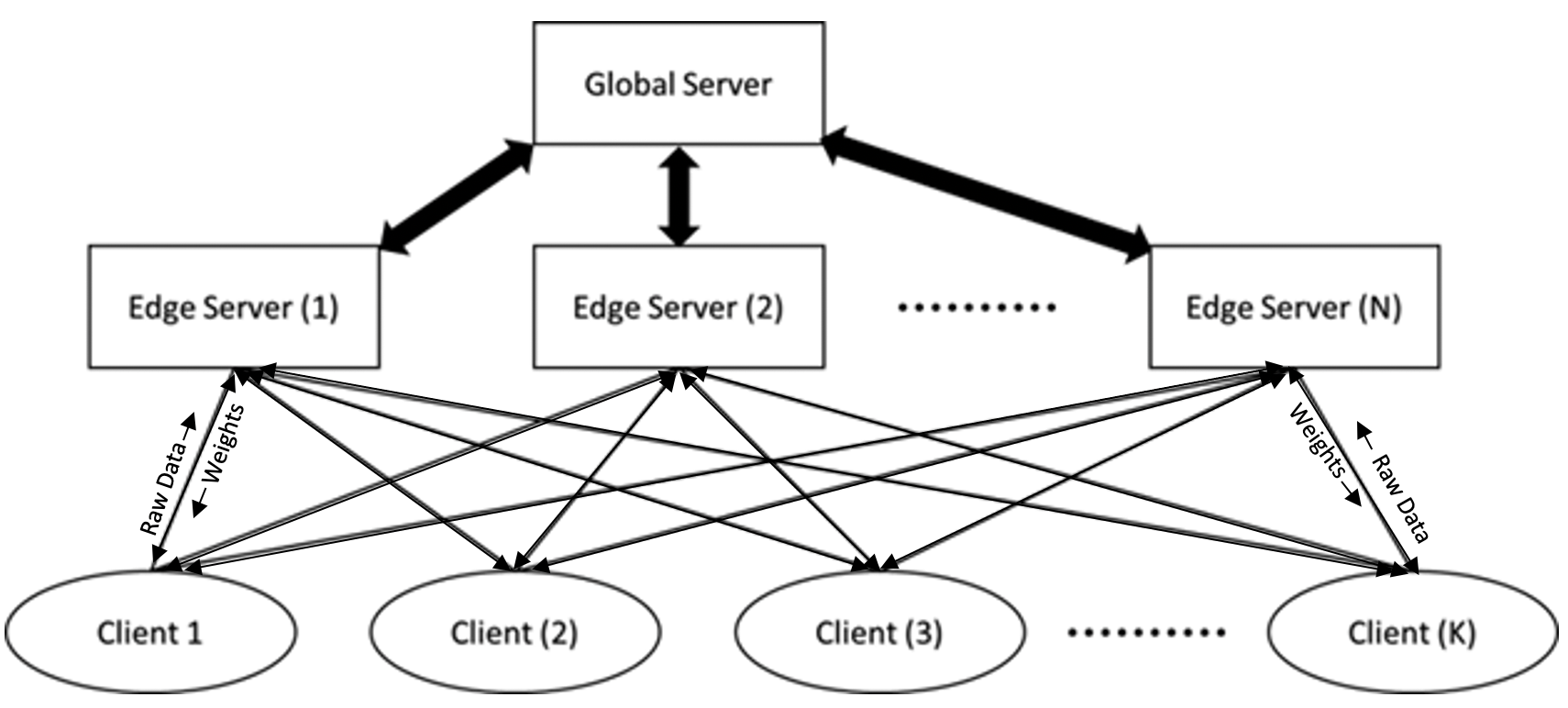}}
\caption{Hierarchical Federated Learning Architecture (in the flow, the "raw data" is network traffic, whereas the "weights" are learned parameters).}
\label{fig1}
\end{figure}

\subsection{Hierarchical Federated Learning  Architecture}
Our HFL environment has three main components: clients, edge servers, and a global server. In the following, we discuss each component briefly. Fig.1 shows the overall architecture of HFL architecture.

\smallskip
\textbf{1. Client:}
The client is a device/machine that owns the data. Each client's data does not move directly from one client to another or even to the edge or global server to preserve privacy. Each client also has local models (independent copies of the same DL models at the client level). Every client can connect to one or multiple edge servers based on availability. The local model on each client gets trained for one or a few epochs, and all clients' updated parameters (trainable weights) of local models are passed to the connected edge servers. The edge servers aggregate the parameters received from the clients and pass them to the global server, where it aggregates all the parameters from edge servers and updates the global model. We then evaluate the global model, which is passed back to the client, where the parameters of the local model are updated. We can have any number of clients in our architecture. Let us assume we have $i$ number of clients symbolized as $c_i$ and each client has its data $X_{c_i}$ and a local model $f_{c_i}$, where $X_{c_i} \in \mathbf{R}^{n\times d}$, where $n$ is the number of samples, and $d$ shows the number of features in each sample. After training $f_{c_i}$ on $m_e$ epochs on $X_{c_i}$ data we pass updated parameters $w_{f_{c_i}}$ of local $f_{c_i}$ model to the connected edge server $e_i$.

\smallskip
\textbf{2. Edge server:}
The edge server is an intermediate between the client and the global server. The edge server is used to add another level of distribution in FL to reduce the load of the global server. The updated weights are passed to the edge server after a few training epochs on local models on client devices. On edge servers, those parameters are aggregated using the selected aggregation function. The aggregation helps reduce the global server load as it has to aggregate thousands or millions of clients, but now the load is distributed among edge servers. We can have any number of edge servers in our architecture. Assume an edge server $e_i$ receives parameters from $m_i$ connected clients; it will aggregate all those parameters by taking the average as given in equation \eqref{eq1} and pass those aggregated parameters $w_{ei}$ to global server $g$.
\begin{equation}\label{eq1}
w_{ei} = \frac{1}{m_i} \sum_{j = 1}^{m_i} w_{cj}
\end{equation}
	
\smallskip
\textbf{3. Global server:}	
The global server resides where the global model resides, and all edge servers are connected to the global server. The parameters of the global model are relayed to all clients after the performance evaluation of the global model. Aggregated parameters on each edge server are passed to the global server, where they are again aggregated based on the same aggregation function. Let us assume global server $g$ is connected to $k$ edge servers; the global server will aggregate all aggregated parameters that it received from edge servers as $w_g$ as shown in equation \eqref{eq2} and will update the parameters of its global model $h$. 
	
	\begin{equation}\label{eq2}
	    w_{g} = \frac{1}{K} \sum_{i = 1}^{k} w_{mi} 
	\end{equation}

\begin{algorithm}
 \label{alg1}
 \caption{HFL Training}
 \begin{algorithmic}[1]
 \renewcommand{\algorithmicrequire}{\textbf{Input:}}

 \REQUIRE training data $X_c$ on each client 
\\ Procedure:
\STATE Initiate local and global models 
\STATE Load and preprocess data on local devices
\FOR {$i = 0$ to $itrs$}
  \STATE get parameters of global models and update local models on clients
  \STATE Train local models on clients for $e$ epochs  
  \STATE Find and connect available edge server and pass parameters to edge server 
  \STATE Aggregate parameters from all clients on edge server
  \STATE Pass aggregated parameters to global server
  \STATE Aggregate parameters from all edge servers and update parameters of global model
  \STATE Evaluate performance of global model
  \ENDFOR
 \RETURN parameters 
 \end{algorithmic} 
 \end{algorithm}

\textbf{Termination condition/epochs in Alg. 1:} The number of epochs will determine how frequently the weights in the neural network will be updated. In order to increase the generalization capabilities of the neural network, the training should occur on an optimal number of epochs.

In general, there is no fixed number of epochs for improving a model's performance. The number of epochs is not specifically relevant as long as a desired accuracy or performance can be achieved. Moreover, the training and validation loss (i.e., errors) better determine the termination and progress of training. As long as these losses decrease, learning happens successfully.

We use `early stopping' to determine when to stop the training. The early stopping method is used to overcome overfitting, where the model performs well on the training data but not on test data. We pass the validation data into the fit() method while fitting our model to the training data. The early stopping technique helps by defining a large number of epochs to train the model and halts the training once the performance stops improving on the validation data.

\subsection{Model Architecture}

\begin{figure}[h!]
\centerline{\includegraphics[width=0.25\textwidth]{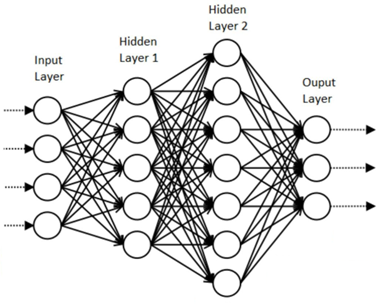}}
\caption{Artificial Neural Networks}
\label{fig2}
\end{figure}

Our system uses fully-connected ANN as local and global models. ANN is a powerful model inspired by the human brain's architecture. ANNs are widely used for supervised learning tasks involving network anomaly detection. We train an ANN on our tabular data. A perceptron is a basic unit in an ANN that takes the dot product of data/input with a weight matrix. The weight matrix contains trainable parameters of the perceptron that learns the basic patterns in data. There can be multiple hidden layers in an ANN, and each layer may have multiple perceptrons. After each layer, a non-linear activation function is used for different non-linearly separable classes. In our model, we use two hidden layers; in each hidden layer, we use 256 hidden units (perceptrons). After each hidden layer, we use ReLU as an activation function ($\sigma$).
In contrast, at the output layer of an ANN, we use a sigmoid activation function that predicts the probabilities of different classes for the given sample. Fig.2 shows the architecture of the ANN model used in our experiments. Let us assume a data sample $X_i$ is passed to the $l^{th}$ layer of the model here $X_i \in \mathbf{R}^{1\times d}$, where $d$ shows the number of features in the given sample the model will work as follows:
\begin{equation}
    o_l = \sigma (w_l^T . X_i + b_l).
\end{equation}

\section{Experimentation and evaluation}
\subsection{Data background}

    In this research, we used the UNSW-NB15 dataset \cite{b1}, and all experiments were conducted with TensorFlow Federated (TFF) in Google Colab with  NVIDIA Tesla P100 and 16 GB RAM. UNSW-NB15 is a network intrusion dataset containing records from raw network packets. The raw network packets of the UNSW-NB 15 dataset were created by the IXIA PerfectStorm tool in the Cyber Range Lab of UNSW Canberra, Australia, for generating a hybrid of real modern normal activities and synthetic recent attack behaviors. The tcpdump tool captured 100 GB of the raw traffic (e.g., Pcap files). This dataset has nine types of attacks: Fuzzers, Analysis, Backdoors, DoS, Exploits, Generic, Reconnaissance, Shellcode, and Worms. The Argus and Bro-IDS tools are used, and twelve algorithms are developed to generate 49 features \cite{b1} with the class label. We used this tabular dataset to detect anomalous network traffic. The dataset contains 2,540,043 samples, and the label column is binary, showing an attack/anomaly if a label is one; otherwise, 0.

\subsubsection{Preprocessing of our data set}
In our experiments, we dropped the null value columns during the preprocessing stage for columns containing null values in more than 50\% samples. After that, we ended up with 45 features, excluding the last four features in \cite{b1}. Then we converted the categorical data into numerical data. We split our data into training and testing data based on a 90-10 ratio and ended with 2,286,038 training and 254,005 testing samples. We used all testing data for the global model evaluation, but we split training data into 4 splits (as we had four clients, so one split for each client), resulting in an equal number of samples. In each training split for each client, we had 571,509 samples, with an 87.5: 12.5 ratio of positive and negative samples in each split.

\subsection{Performance and evaluation metrics}

Performance and evaluation metrics are used to evaluate the performance of ML models. Mostly in supervised learning techniques involving network anomaly detection, the model's performance is evaluated by comparing the model's prediction with the actual ground truth. The evaluation metrics for our experiments include accuracy, precision, recall, and F1-Score. 

Precision represents how precise/accurate each model is out of those data points that are predicted as anomalous and how many of them are actually anomalous. A high precision (seen in our deep neural network model) is a good measure considering an unbalanced data set and is robust against false positives. (False positives occur when a model mispredicts the positive outcome. An incorrectly predicted negative class is referred to as a false negative). The recall represents how many of the actual network anomalies the models captured through predicting it as anomalous. F1-score is a balance of both precision and recall, given as $F1\: score=2\times\frac{precision \times recall}{precision + recall}$. Thus, it ensures that high precision or recall alone cannot bias the metric. Finally, accuracy is the simple mean of correctness derived from the difference in predictions from the labeled ground truth data.

\begin{figure}[h]
\centering
\includegraphics[width=0.33\textwidth]{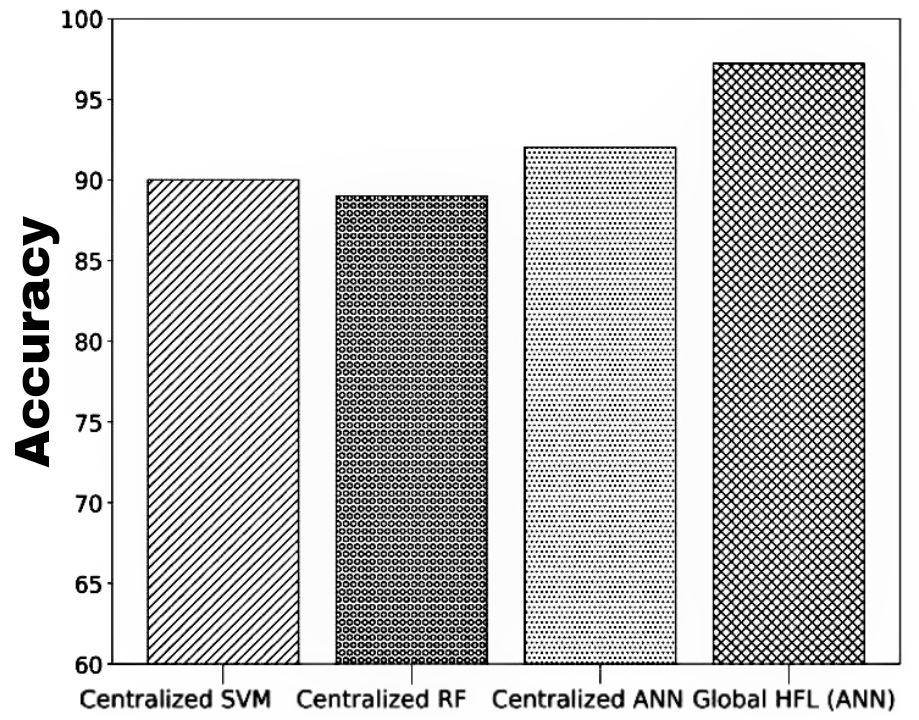}
\caption{Accuracy of ML techniques (network anomaly detection)}
\label{fig:acuracy-comparison}
\end{figure}

\subsection{Performance evaluation}

This sub-section discusses the performance evaluation results on different models and experimental settings. Here we can see that the performance of local models improves when they are trained in an HFL environment. In Fig.3, we have shown that the ANN model trained in HFL has the best performance with an accuracy of 97.21\% and a training time of 1,500 seconds (please see Fig.5). Even though the performance of ANN trained in a traditional centralized  ML environment came close, it took longer to train under the same conditions. Support vector machine (SVM) and random forests models had accuracies of 90.01\% and 89.01\%, respectively, in the centralized ML approaches. Furthermore, Table I shows different experiments' evaluation scores for accuracy, precision, recall, and F1-scores. An illustrative comparison is shown in Fig. 4.

\begin{figure}[h]
\centerline{\includegraphics[width=0.33\textwidth]{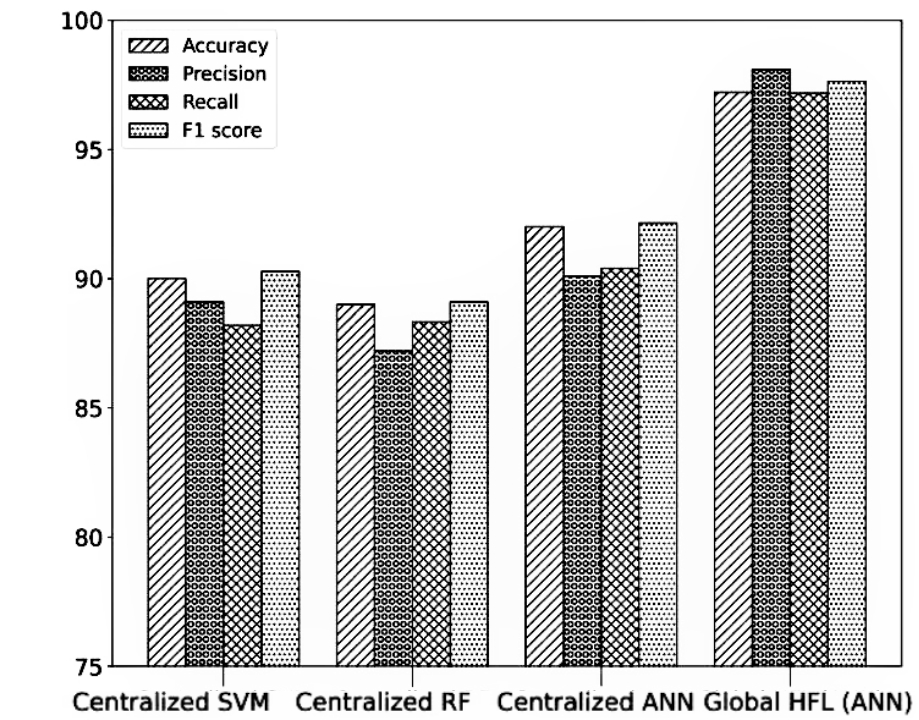}}
\caption{ML algorithms performance measures for UNSW-NB15 dataset.}
\label{fig:acuracy-comparison2}
\end{figure}

\begin{figure}[h]
\centerline{\includegraphics[width=0.28\textwidth]{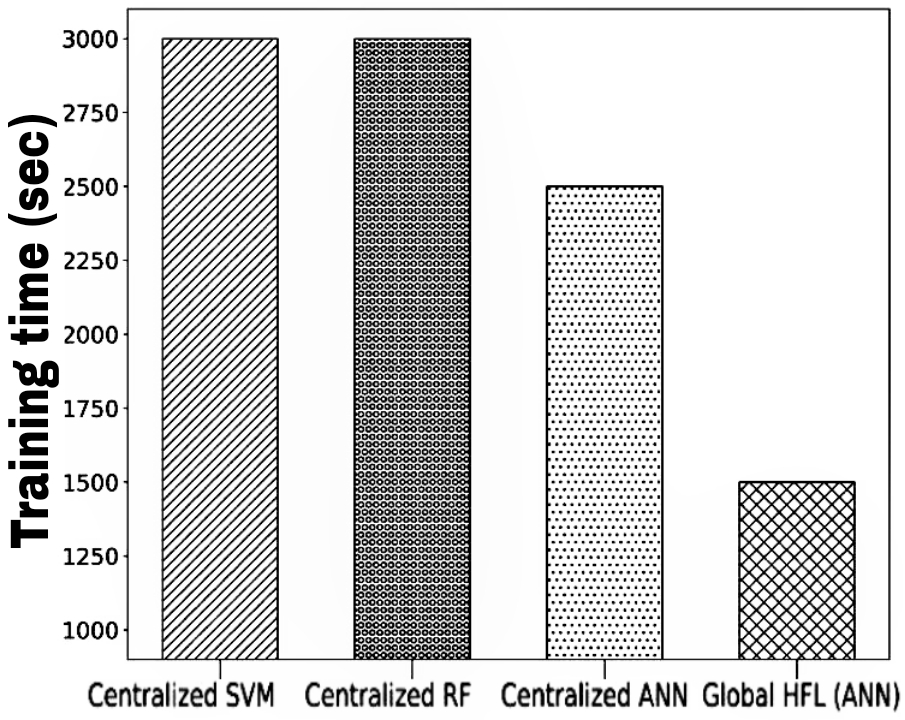}}
\caption{Training time cost of ML algorithms}
\end{figure}

\subsection{Experimental results}

This section discusses other experiments we performed to develop an efficient and accurate network anomaly detection algorithm. In this work, we performed the following types of experiments:

\smallskip
\textbf{1. Experiments of model selection:}

We used all the data, including training and testing splits of the data set, to select the suitable ML model. For this purpose, we compared the performance of SVM, Random Forest, and ANN on the test data. Table I shows that the ANNs performed well compared to SVM and Random Forest in these experiments; that is why we used ANNs in further experiments. It further reinforces the higher accuracy gains in HFL over individual clients.

\smallskip
\textbf{2. Network anomaly detection on individual clients:}

In these experiments, we divided our data set into 4 parts with the same positive and negative sample ratio. These splits of the data set have different distributions of different types of attacks. We then trained ANN models for each data split separately and evaluated the performance using selected evaluation metrics. In Table I, we can see that the performance of different individual models trained with different splits of data varies (for example, on client/split 1, accuracy is about 90\% whereas, on client 3, it is about 91\%). We can also see that the overall performance of individually trained models is inferior to those of the same data splits when models were trained in the FL environment. From the learning curves in Fig.6, we can see that initially, the learning of all models was slow (e.g., iterations 1 through 5), but after several iterations (please see Fig.6), it speeds up as \textbf{global models start to learn patterns} learned by the local models from local data.

\begin{table}[h]
\label{tab:acuracy-comparison}
\vspace*{1mm}
\smallskip
\caption{Performance evaluation}
\begin{tabular}{|l|l|l|l|l|}
\hline
\textbf{Experiment}   & Accuracy & Precision & Recall & F1 Score                      \\ \hline
{Client 1 (Individual)} & 90.12 & 91.89 & 91.42 & 91.65  \\ \hline
{Client 1 (HFL)}         & 96.41 & 98.46 & 97.65 & 98.05 \\ \hline
{Client 2 (Individual)} & 90.98 & 91.63 & 92.03 & 91.82 \\ \hline
{Client 2 (HFL)}         & 96.39 & 98.5  & 97.82 & 98.15 \\ \hline
{Client 3 (Individual)} & 91.01 & 91.58 & 92.07 & 91.82 \\ \hline
{Client 3 (HFL)}         & 96.58 & 98.58 & 97.79 & 98.18 \\ \hline
{Client 4 (Individual)} & 90.92 & 92.03 & 91.51 & 91.76 \\ \hline
{Client 4 (HFL)}         & 96.42 & 98.57 & 97.8  & 98.18 \\ \hline
Centralized SVM             & 90.01        & 89.10         & 88.19      & 90.28                      \\ \hline
Centralized RF  & 89.01        & 87.21         & 88.32      & 89.11                      \\ \hline
Centralized ANN           & 92.01        & 90.10         & 90.40      & 92.15                      \\ \hline
Global HFL (ANN) \textbf{(ours)}      & 97.21   & 98.09     & 97.18  & 97.63                     \\ \hline
\end{tabular}
\end{table}

\begin{table}[h]
\label{tab:acuracy-comparison}
\vspace*{1mm}
\smallskip
\caption{Performance evaluation}
\begin{tabular}{|l|l|l|l|l|}
\hline
\textbf{Experiment}   & Accuracy & Precision & Recall & F1 Score                      \\ \hline
{4 Clients / 4 Edge Servers}  & 96.02 & 98.12 & 97.32 & 97.71 \\ \hline
{6 Clients / 3 Edge Servers}  & 96.19 & 98.25 & 97.52 & 97.88 \\ \hline
{10 Clients / 3 Edge Servers} & 96.39 & 98.53  & 97.81 & 98.15 \\ \hline
{10 Clients / 4 Edge Servers} & 96.37 & 98.49 & 97.82  & 98.14 \\ \hline
{10 Clients / 5 Edge Servers} & 97.38 & 98.51 & 97.81 & 98.15 \\ \hline
\end{tabular}
\end{table}


\begin{figure}[h]
\centerline{\includegraphics[width=0.46\textwidth]{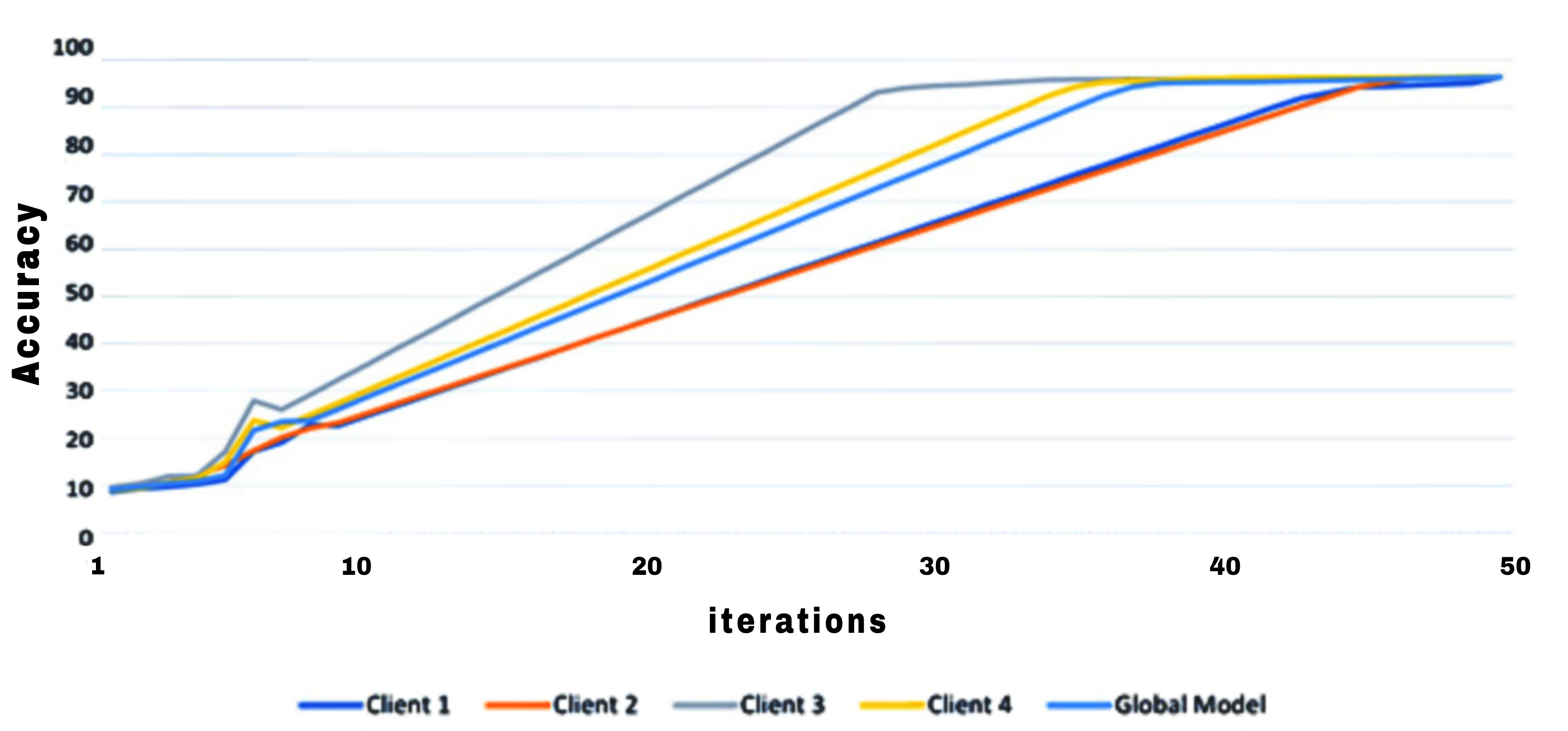}}
\caption{Learning curve of the local and global models in 50 iterations.}
\label{fig:acuracy-comparison2}
\end{figure}


\textbf{3. Network anomaly detection in FL environment:}

In these experiments, we explored the effect on the performance of the network anomaly detection models if we increase or decrease the number of clients. We experimented using 4, 6, and 10 clients, keeping the count of edge servers constant (3), and then we changed the number of edge servers to 4 and 5, keeping the number of clients constant (10). In these experiments, we first created 10 splits of the dataset, and when we used only $s$ number of splits for experiments having the same number of clients. The results show that the edge servers increase the federated learning architectures' scalability while retaining the model's performance in the system. 

Notably, aggregations are done first on edge servers before passing them to the global server; this reduces the load on the global server and makes communication faster. In literature, trade-offs have been reported between the communication cost and training time depending on the communication frequency with the servers, e.g., in the investigation presented in \cite{16}. We delegate a further investigation on bandwidth vs. accuracy/efficiency as future work.

In our experiments, the highest accuracy we achieved with 10 clients was 97.38\%; meanwhile, the training time more than doubled from that of 4 clients. Table II shows the performance of global models trained on a different number of clients and edge servers. We can see that increasing the number of clients enhances the performance of the models; however, there is no significant impact on performance if we change the number of edge servers, but the training time increases, on the other hand.

Another salient point is that centralized ANN and HFL had similar performances in terms of accuracy. However, the only difference was the training time, as the HFL trained much faster due to the inclusion of the edge servers that reduced the load on the global model. Therefore, with time kept constant, the HFL method could train itself much more than centralized ANN. This explains the accuracy gains of the HFL method besides only efficiency.

\section{Conclusion}
In contemporary times, not only computers or mobile devices but also many IoT-based home appliances are connected to private or public networks. Besides many advantages, there is always a threat of network attacks on the devices connected to networks, especially the internet. In this work, we proposed to use ANN in an HFL environment to detect if the requests from a device are anomalous or not. 

HFL helps with scalability issues, makes global models more efficient, and reduces the training load on global servers. Results of our experiments on the UNSW-NB15 dataset have shown the improvement in the performance of ANNs on client devices to detect anomalous requests when trained in the federated learning environment in comparison to when they were trained on the data of individual devices as it only contains the information about attacks they were being targeted. In future work, we will hone in on optimizing the data imbalance problem inherent to biased datasets to further improve the accuracy, and propose to use pre-selection for faster training.

\vspace{12pt}
\end{document}